\documentclass{article}

\usepackage{microtype}
\usepackage{graphicx}
\usepackage{booktabs} 

\usepackage{hyperref}

\usepackage[accepted]{icml2020}

\icmltitlerunning{Scene Graph Reasoning for Visual Question Answering}

\usepackage{graphicx}
\usepackage{subcaption}
\usepackage[export]{adjustbox}
\usepackage{amsmath, amssymb}
\usepackage{dsfont}
\DeclareMathOperator*{\argmax}{arg\,max}

\begin{document}

\twocolumn[
\icmltitle{Scene Graph Reasoning for Visual Question Answering}



\icmlsetsymbol{equal}{*}

\begin{icmlauthorlist}
\icmlauthor{Marcel Hildebrandt}{equal,siemens,lmu}
\icmlauthor{Hang Li}{equal,siemens,tum}
\icmlauthor{Rajat Koner}{equal,lmu}
\icmlauthor{Volker Tresp}{siemens,lmu}
\icmlauthor{Stephan G\"unnemann}{tum}
\end{icmlauthorlist}

\icmlaffiliation{siemens}{Siemens AG, Munich, Germany}
\icmlaffiliation{lmu}{Ludwig Maximilian University of Munich, Munich, Germany}
\icmlaffiliation{tum}{Technical University of Munich, Munich Germany}

\icmlcorrespondingauthor{Marcel Hildebrandt}{marcel.hildebrandt@siemens.com}

\icmlkeywords{Visual question answering, Scene graphs, Graph learning, Reinforcement learning}

\vskip 0.3in
]



\printAffiliationsAndNotice{\icmlEqualContribution} 

\begin{abstract}
Visual question answering is concerned with answering free-form questions about an image. Since it requires a deep linguistic understanding of the question and the ability to associate it with various objects that are present in the image, it is an ambitious task and requires techniques from both computer vision and natural language processing. We propose a novel method that approaches the task by performing context-driven, sequential reasoning based on the objects and their semantic and spatial relationships present in the scene. As a first step, we derive a scene graph which describes the objects in the image, as well as their attributes and their mutual relationships. A reinforcement agent then learns to autonomously navigate over the extracted scene graph to generate paths, which are then the basis for deriving  answers. We conduct a first experimental study on the challenging GQA dataset with manually curated scene graphs, where our method almost reaches the level of human performance.
\end{abstract}

\section{Introduction}
\label{sec:introduction}
\begin{figure*}[ht]
    \begin{subfigure}{0.52\textwidth}
        \includegraphics[width=\linewidth, height=6.5cm,
        center]{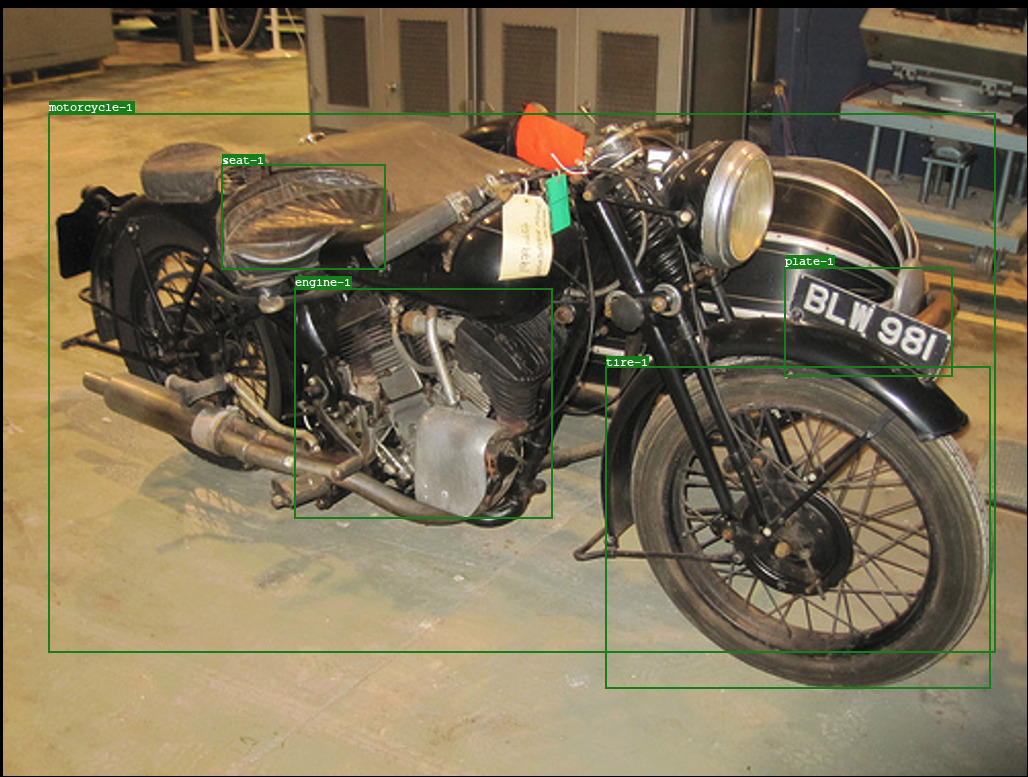} 
        \label{fig:subim1}
    \end{subfigure}
    \begin{subfigure}{0.45\textwidth}
    \vspace{-3mm}
        \includegraphics[width=\linewidth,center]{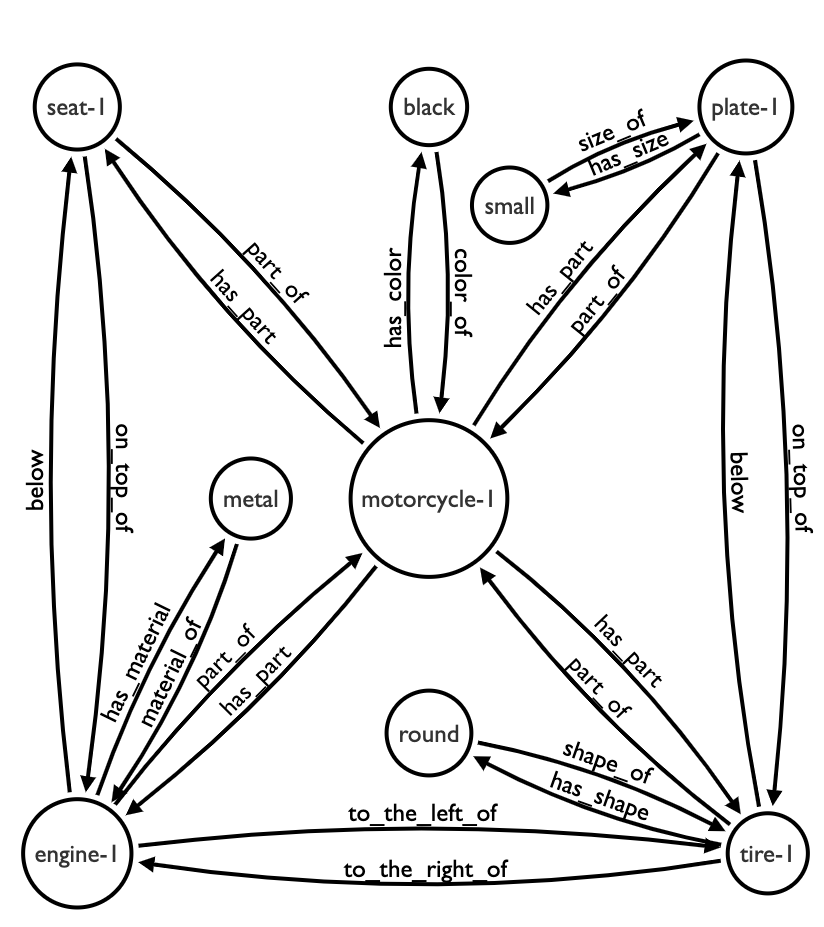}
        \label{fig:subim2}
    \end{subfigure}
    \caption{Example of an image and the corresponding scene graph.
    }
    \label{fig:example_image_sg}
\end{figure*}
Visual Question Answering (VQA) is a demanding task that involves understanding and reasoning over two data modalities: images and natural language. Given an image and a free-form question---which formulates a question about the presented scene--- the issue is for the algorithm to find the correct answer.

VQA has recently found interest in  different research communities and various real-world data sets, such as the \textit{VQA} data set \cite{antol2015vqa},  have been generated. It has been argued that, in the \textit{VQA} data set, many of the apparently challenging reasoning tasks can be solved by an algorithm by exploiting trivial prior knowledge,  and thus by shortcuts to proper reasoning
(e.g., clouds are white or doors are made of wood). 
To address these shortcomings,  the \textit{GQA}  dataset \cite{hudson2019gqa} has been developed.
Compared to other real-world datasets, \textit{GQA}  is more suitable to evaluate reasoning abilities since the images and questions are carefully filtered to make the data less prone to biases. 

Plenty of VQA approaches are agnostic towards the explicit relational structure of the objects in the presented scene and rely on monolithic neural network architectures that process regional features of the image separately \cite{anderson2018bottom, yang2016stacked}. While these methods led to promising results on previous datasets, they lack explicit compositional reasoning abilities which results in weaker performance on more challenging datasets such as \textit{GQA} . 
Other works \cite{teney2017graph, shi2019explainable, hudson2019learning} perform reasoning on explicitly detected objects and interactive semantic and spatial relationships among them. These approaches are closely related to the scene graph representations \cite{johnson2015image} of an image, where detected objects are labeled as nodes and relationship between the objects are labeled as edges.


In this work we aim to combine VQA techniques with recent research advances in the area of statistical relation learning on knowledge graphs (KGs).
KGs provide human readable, structured representations of knowledge about the real world via collections of factual statements. Inspired by multi-hop reasoning methods on KGs such as \cite{minerva, hildebrandt2020reasoning}, we model the VQA task as a path-finding problem on scene graphs. The underlying idea can be summarized with the phrase: Learn to walk to the correct answer. More specifically, given an image, we consider a scene graph and train a reinforcement learning agent to conduct a policy-guided random walk on the scene graph until a conclusive inference path is obtained. In contrast to purely embedding-based approaches, our method provides explicit reasoning chains that leads to the derived answers. To sum up, our major contributions are as follows.
\begin{itemize}
    \item To the best of our knowledge, we propose the first VQA method that employs reinforcement learning for reasoning on scene graphs.
    \item We conduct an experimental study to analyze the reasoning capabilities of our method. Instead of generating our own scene graphs, we consider manually curated scene graphs from the \textit{GQA} dataset for these first experiments. This setting allows to isolate the noise associated to the visual perception task and focuses solely on the language understanding and reasoning task. Thereby, we can show that our method achieves human-like performance.
\end{itemize}

\section{Method}
\label{sec:our_method}

\begin{figure*}
 \begin{center}
    \includegraphics[width=0.8\textwidth]{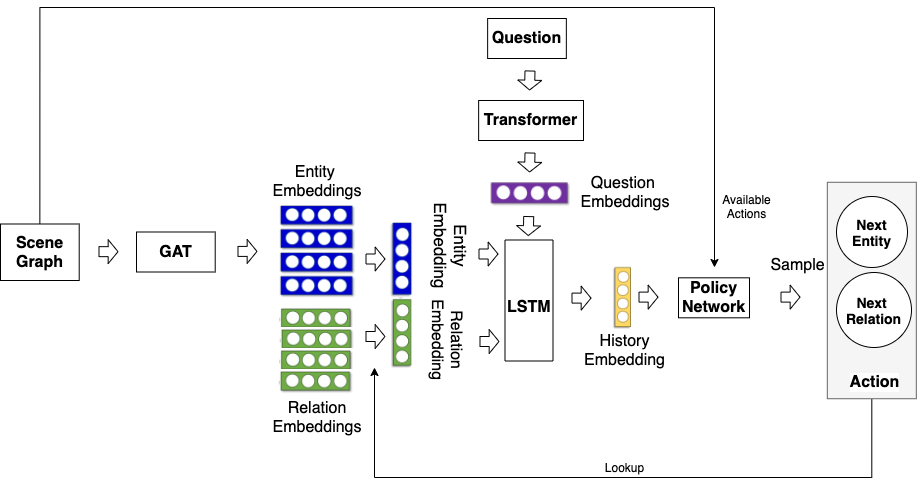}
  \caption{The architecture of our scene graph reasoning module. 
  }
  \label{fig:architecture}   
 \end{center}
\end{figure*}

The task of VQA is framed as a scene graph traversal problem. Starting from a hub node that is connected to all other nodes, an agent sequentially samples transition to a neighboring node on the scene graph until the node corresponding to the answer is reached. In this way,  by adding transitions  to the current path, the reasoning chain is successively extended. Before describing the decision problem of the agent, we introduce the notation that we use throughout this work.

\paragraph{Notation}
A scene graph is a directed multigraph where each node corresponds to a scene entity which is either an object associated with a bounding box or an attribute of an object. Each scene entity comes with a type that corresponds to the predicted object or attribute label. Typed edges specify how  scene entities are related to each other. More formally, let $\mathcal{E}$ denote the set of scene entities and consider the set of binary relations $\mathcal{R}$. Then a scene graph $\mathcal{SG} \subset \mathcal{E} \times \mathcal{R} \times \mathcal{E} $ is a collection of ordered triples $(s, p, o)$ – subject, predicate, and object. For example, as shown in Figure \ref{fig:example_image_sg}, the triple \textit{(motorcycle-1, has\_part, tire-1)} indicates that both a motorcycle (subject) and a tire (object) are detected in the image. The predicate \textit{has\_part} indicates the relation between the entities. Moreover, we denote with $p^{-1}$ the inverse relation corresponding to the predicate $p$. For the remainder of this work, we impose completeness with respect to inverse relations in the sense that for every $(s, p, o) \in \mathcal{SG}$ it is implied that $(o, p^{-1}, s) \in \mathcal{SG}$. Moreover, we add a so-called hub node (\textit{hub}) to every scene graph which is connected to all other nodes. 

\paragraph{Environment}\sloppy

The state space of the agent $\mathcal{S}$ is given by $\mathcal{E} \times \mathcal{Q}$ where $\mathcal{E}$ are the nodes of a scene graph $\mathcal{SG}$ and $\mathcal{Q}$ denotes the set of all questions. The state at time $t$ is the entity $e_t$ at which the agent is currently located and the question $Q$. Thus, a state $S_t \in \mathcal{S}$ for time $t \in \mathbb{N}$ is represented by $S_t = \left(e_t, Q\right)$. The set of available actions from a state $S_t$ is denoted by $\mathcal{A}_{S_t}$. 
It contains all outgoing edges from the node $e_t$ together with their  corresponding object nodes. 
More formally, $\mathcal{A}_{S_t} = \left\{(r,e) \in \mathcal{R} \times \mathcal{E} :  S_t = \left(e_t, Q\right) \land \left(e_t,r,e\right) \in \mathcal{SG}\right\}\, .$ Moreover, we denote with $A_t \in \mathcal{A}_{S_t}$ the action that the agent performed at time $t$. We include self-loops for each node in $\mathcal{SG}$ that produce  a \textit{NO\_OP}-label. These self-loops  allow the agent to remain at the current location if it reaches the answer node. To answer binary questions, we also include artificial \textit{yes} and \textit{no} nodes in the scene graph. The agent can transition to these nodes in the final step.
 

\paragraph{Embeddings} We initialize words in $Q$ with  GloVe embeddings \cite{pennington2014glove} with dimension $d=300$. Similarly we initialize entities and relations  in $\mathcal{SG}$   with the embeddings of  their type labels.
In the scene graph, the node embeddings are passed through a multi-layered graph attention network (GAT) \cite{velivckovic2017graph}. Extending the idea from graph convolutional networks \cite{kipf2016semi} with a  self-attention mechanism, GATs mimic the convolution operator on regular grids where an entity embedding is formed by aggregating node features from its neighbors. Thus, the resulting embeddings are context-aware, which makes nodes with the same type but different graph neighborhoods distinguishable.  To produce an embedding for the question $Q$, we first apply a Transformer \cite{vaswani2017attention},  followed by a mean pooling operation.

\paragraph{Policy}
We denote the agent's history until time $t$ with the tuple $H_t = \left(H_{t-1}, A_{t-1}\right)$ for $t \geq 1$ and $H_0 = hub$ along with $A_0 = \emptyset$ for $t = 0$. The  history is encoded via a multilayered LSTM \cite{hochreiter1997long}
\begin{equation}
\label{eq:lstm_agent}
        \mathbf{h}_t = \textrm{LSTM}\left(\mathbf{a}_{t-1}\right) \, ,
\end{equation}
where $\mathbf{a}_{t-1} = \left[\mathbf{r}_{t-1},\mathbf{e}_{t}\right] \in \mathbb{R}^{2d}$ corresponds to the embedding of the previous action with $\mathbf{r}_{t-1}$ and $\mathbf{e}_{t}$ denoting the embeddings of the edge and the target node into $\mathbb{R}^{d}$, respectively. 
The history-dependent action distribution is given by
\begin{equation}
\label{eq:policy_agent}
\mathbf{d}_t = \textrm{softmax}\left(\mathbf{A}_t \left(\mathbf{W}_2\textrm{ReLU}\left(\mathbf{W}_1 \left[ \mathbf{h}_t, \mathbf{Q} \right]\right)\right)\right) \, ,
\end{equation}
where the rows of $\mathbf{A}_t \in \mathbb{R}^{\vert \mathcal{A}_{S_t} \vert \times d}$ contain latent representations of all admissible actions. Moreover, $\mathbf{Q} \in \mathbb{R}^{d}$ encodes the question $Q$. The action $A_t = (r,e) \in \mathcal{A}_{S_t}$ is drawn according to $\textrm{categorical}\left(\mathbf{d}_t\right)$.
Equations \eqref{eq:lstm_agent} and \eqref{eq:policy_agent} induce  a stochastic policy $\pi_{\theta}$, where $\theta$ denotes the set of trainable parameters.

\begin{table*}[ht!]
\caption{A comparison of our method with human performance based on manually curated scene graphs.
}
\begin{center}
\resizebox{0.8\textwidth}{!}{
\begin{tabular}{c c c c c c c c}
\hline
Method & Binary & Open & Consistency & Validity & Plausibility & Accuracy \\
\hline
Human \cite{hudson2019gqa}   & 91.2 & 87.4 & 98.4 & 98.9 & 97.2 &  89.3 \\
TRRNet & 	77.91&	50.22&	89.84&	85.15&	96.47&	63.20 \\
Our Method    & 90.41 & 90.86 & 91.92 & 93.68 & 93.13  & 90.63  \\
\hline
\end{tabular}
}
\label{tab:gt_results}
\end{center}
\end{table*}

\begin{figure*}
    \begin{subfigure}{0.315\textwidth}
        \includegraphics[width=\linewidth,center]{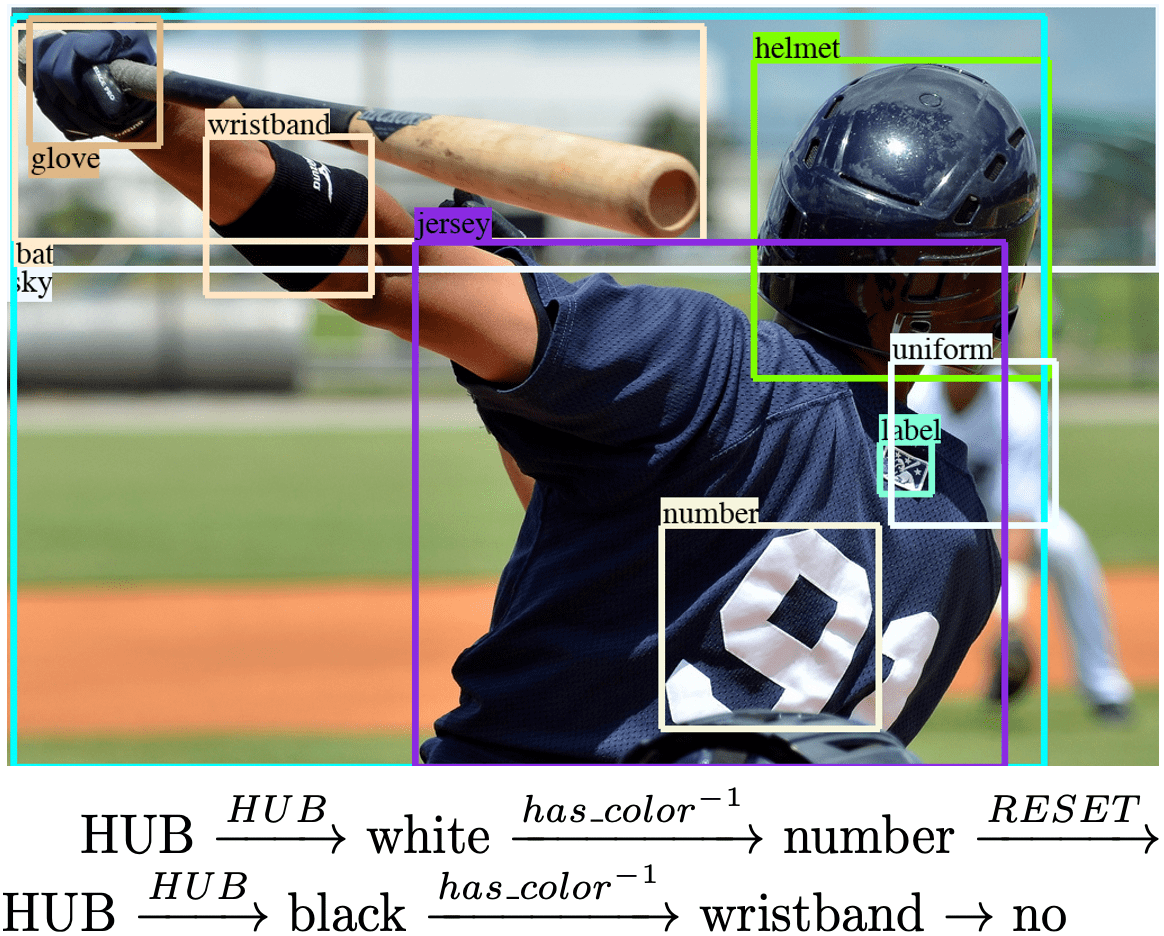} 
        \caption{Question: Is the color of the number the same as that of the wristband? \\Answer: No.}
        \label{subfig:VQA_ex1}
    \end{subfigure}
    \hspace{1mm}
    \begin{subfigure}{0.31\textwidth}
        \includegraphics[width=\linewidth,center]{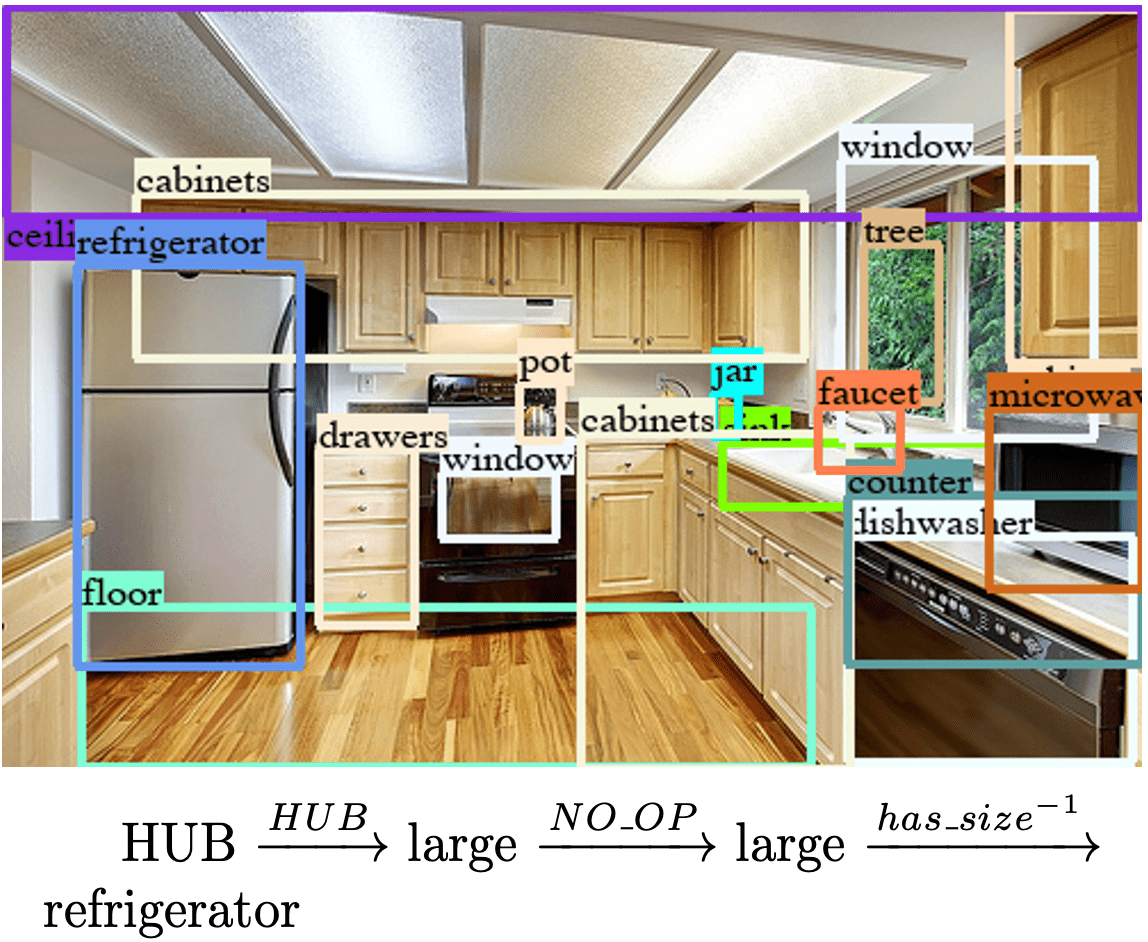}
        \caption{Question: What is the name of the appliance that is not small? \\Answer: Refrigerator.
}
        \label{subfig:VQA_ex2}
    \end{subfigure}
    \hspace{1mm}
    \begin{subfigure}{0.31\textwidth}
        \includegraphics[width=\linewidth,center]{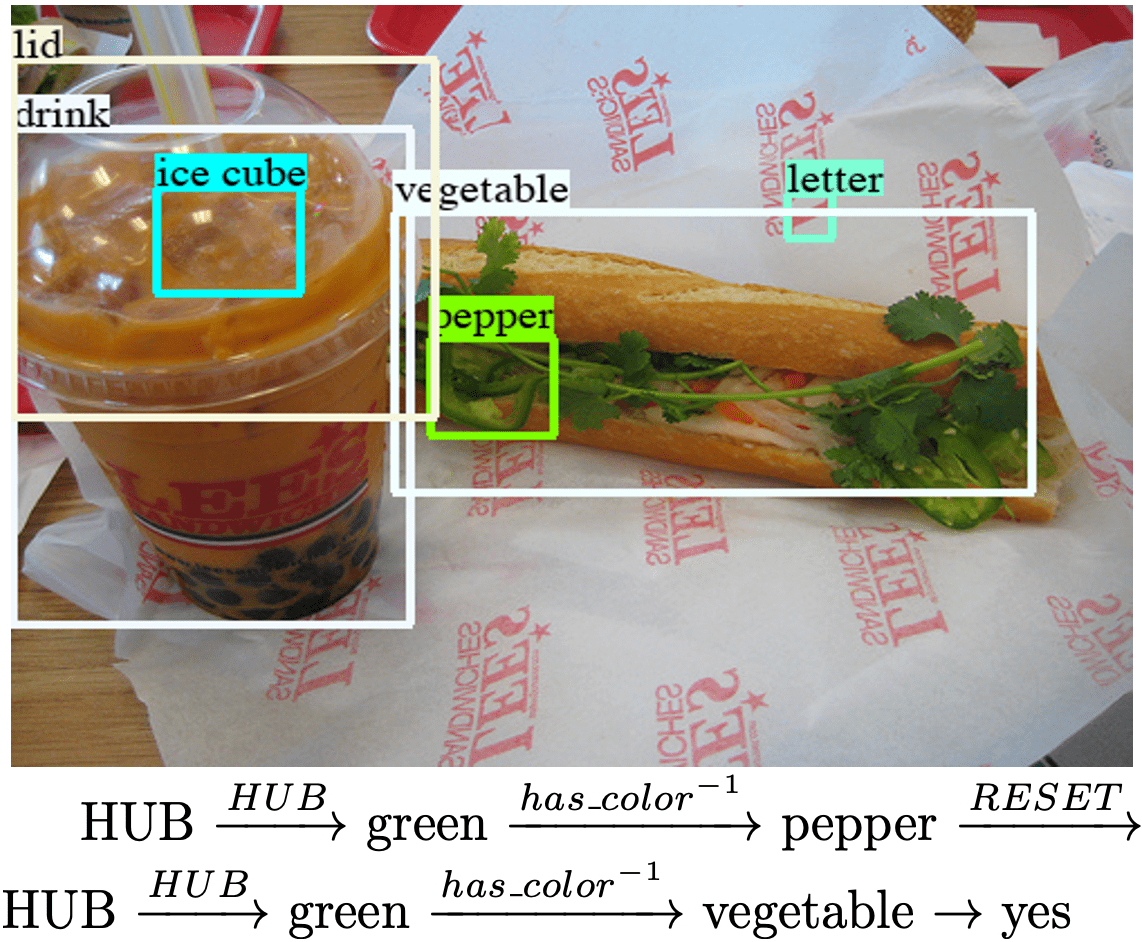}
        \caption{Do both the pepper and the vegetable to the right of the ice cube have green color? \\Answer: Yes.
}
        \label{subfig:VQA_ex3}
    \end{subfigure}
    \caption{Three examples question and the corresponding images and paths.}
    \label{fig:example_paths}
\end{figure*}

\paragraph{Rewards and Optimization}
After sampling $T$ transitions, a terminal reward is assigned according to
\begin{equation}
    R = \begin{cases}
 1 &\text{if $e_T$ is the answer to $Q$,} \\
0 &\text{otherwise.}
\end{cases}
\end{equation}
We employ REINFORCE \cite{williams1992simple} to maximize the expected rewards. Thus, the agent's maximization problem is given by
\begin{equation}
\label{eq:objective_agent}
    \argmax_{\theta} \mathbb{E}_{Q \sim \mathcal{T}}\mathbb{E}_{A_1, A_2, \dots, A_N \sim \pi_{\theta}}\left[R \left\vert\vphantom{\frac{1}{1}}\right. e_c \right] \, ,
\end{equation}
where $\mathcal{T}$ denote the set of training questions.

\if false
\subsection{Related Work}

Various models have been proposed that perform VQA on both real-world and artificial datasets. Currently, most dominant VQA approaches can be categorized into two different branches: First, monolithic neural networks which perform implicit reasoning on latent representations obtained from fusing the two data modalities. Second,  multi-hop methods that form explicit symbolic reasoning chains on a structured representation of the data. Monolithic network architectures obtain visual features from the image either in from individual detected objects or by processing the whole image directly via convolutional neural networks (CNNs). The derived embeddings are usually scored against a fixed answer set along with the embedding of the question obtained from a sequence model. Moreover, co-attention mechanism are frequently employed to couple the vision and the language models allowing for interactions between objects from both modalities \cite{kim2018bilinear, anderson2018bottom, cadene2019murel, yu2017multi, zhu2017structured}. Monolithic networks are among the dominant methods on previous real-world VQA datasets such as \cite{antol2015vqa}. 
However, they suffer from the black-box problem and posses limited reasoning capabilities with respect to complex questions that require long reasoning chains (see \cite{chen2019meta} for a detailed discussion).

Explicit reasoning methods combine the sub-symbolic representation learning paradigm with symbolic reasoning approaches over structured representations of the image. Most of the popular explicit reasoning approaches follow the idea of neural module networks (NMNs) \cite{andreas2016neural} which perform a sequence of reasoning steps realized by forward passes through specialized neural networks that each correspond to predefined reasoning subtasks. Thereby, NMNs construct functional programs by dynamically assembling the modules resulting in a question-specific neural network architecture. In contrast to the monolithic neural network architectures described above, these method contain a natural transparency mechanism via the functional programs. However, while NMN-related methods (e.g., \cite{hu2017learning, mao2019neuro}) exhibit good performance on synthetic datasets such as CLEVR \cite{johnson2017clevr}, they require functional module layouts as additional supervision signal to obtain good results. Hence, NMNs have a limited applicability in most real-world settings. In this paper, we propose a novel neuro-symbolic scene graph reasoning approach which leverages representation learning combined with explicit reasoning on scene graphs. Closely related to our method is the Neural State Machine (NSM) proposed by \cite{hudson2018compositional}. The underlying idea of NMS consists in first constructing a scene graph from an image and treat it as a state machine. Concretely, the nodes correspond to states and edges to transitions. Then conditioned on the question, a sequence of instructions is derived that indicates how to traverse the scene graph and arrive at the answer. In contrast to NSM we treat path-finding as a decision problem in a reinforcement learning setting. Concretely, we outline in the next section how extracting predictive paths from scene graphs can be naturally formulated in term of a goal-oriented random walk induced by a stochastic policy that allows to balance between exploration and exploitation. Moreover, our framework integrates state-of-the-art techniques from graph representation learning and NLP. In this paper we only consider basic policy gradient methods but more sophisticated reinforcement learning technique lo s can be be employed in future works.


\paragraph{Statistical Relational Learning} Machine learning methods for KG reasoning aim at exploiting statistical regularities in observed connectivity patterns. These methods are studied under the umbrella of statistical relational learning (SRL) \cite{nickel2015review}. In the recent years KG embeddings have become the dominant approach in SRL. The underlying idea is that graph features that explain the connectivity pattern of KGs can be encoded in low-dimensional vector spaces. In the embedding spaces  the interactions among the embeddings for entities and relations can be efficiently modelled to produce scores that predict the validity of a triple. Despite achieving good results in KG reasoning tasks, most embedding based methods hardly capture the compositionality expressed by long reasoning chains. 

Recently, multi-hop reasoning methods such as MINERVA  \cite{minerva} and DeepPath \cite{wenhan_emnlp2017} were proposed. Both methods are based on the idea that a reinforcement learning agent is trained to perform a policy-guided random walk until the answer entity to a query is reached. Thereby, the path finding problem of the agent can be modeled in terms of a sequential decision making task framed as a Markov decision process (MDP). The method that we propose is in this work follow a similar philosophy in the sense that we train an RL agent to navigate on a scene graph to the correct answer node. However, a conceptial difference is that he agents in MINERVA and DeepPath perform walks on large scale knowledge graphs exploiting repeating, statistical patterns. Thereby, the policies implicitly incorporate (fuzzy) rules. In addition, instead of processing free-form questions, the query in the KG reasoning setting is structured as a pair of symbolic entities. That is why we propose a wide range of modification to adjust our method to the challenging VQA setting.

\subsection{Connection to Cognitive Science}
VOLKER

\fi

\section{Experiments}
\label{sec:experiments}

\paragraph{Dataset and Experimental Setup}
 \citet{hudson2019gqa} introduced the \textit{GQA}  dataset with the goal of addressing key shortcomings of previous VQA datasets, such as  \textit{CLEVR}  \cite{johnson2017clevr} or the \textit{VQA} dataset \cite{antol2015vqa}.  \textit{GQA} is more suitable for evaluating  reasoning and compositional abilities of a model,  in a realistic setting. The \textit{GQA}  dataset contains 113K images and around 1.2M questions split into roughly $80\%/10\%/10\%$ for the training, validation and testing. The overall vocabulary size consists of 3097 words.

\if false
All experiments were conducted on a machine with one NVIDIA Tesla K80 GPU and 64 GB RAM. Training the scene graph reasoner of MODELNAME for 20 epochs on \textit{GQA}  takes around 14 hours, testing about 2 hours. 

We compare the performance of our method with the baselines in \cite{hudson2019gqa} as well as with the top single submission to the \textit{GQA}  challenge where we only consider those models that did not use the functional program for additional supervision but only the answers as a supervision signal.
\fi

\paragraph{Results and Discussion}
\label{sec:results}

In this work, we report first results on an experimental study on manually curated scene graphs that are provided in the \textit{GQA}  dataset. In this setting, the true reasoning and language understanding capabilities of our model can be analyzed.  
Table \ref{tab:gt_results} shows the  performance of our method and compares it  with the human performance reported in \cite{hudson2019gqa} and with  TRRNet\footnote{\url{https://evalai.cloudcv.org/web/challenges/challenge-page/225/leaderboard/733\#leaderboardrank-5}}, the best performing single method submission to the \textit{GQA}  Real-World Visual Reasoning Challenge 2019. 
Along with the accuracy on open questions (``Open''), binary questions (yes/no) (``Binary''), and the overall accuracy (``Accuracy''), we also report the additional metric ``Consistency'' (answers should not contradict themselves), ``Validity'' (answers are in the range of a question; e.g., \textit{red} is a valid answer when asked for the color of an object), ``Plausibility'' (answers should be reasonable;  e.g., red is a reasonable color of an apple reasonable, blue is not), as  proposed in \cite{hudson2019gqa}.

The results in Table \ref{tab:gt_results} show that our method achieves human level performance with respect to most metrics. Figure \ref{fig:example_paths} shows three examples of scene graph traversals, which produced the correct answer. 
An advantage of our approach is that the sequential reasoning process makes the model output  transparent and easier to debug in cases of failures. 

Although the results in Table \ref{tab:gt_results} are very encouraging, the performance numbers are not directly comparable, since the underlying data sets are different and since we operated on manually curated scene graphs.
 As part of ongoing work, we are exploring different methods for extracting the scene graph  automatically from the images. This step is not really the focus of this work but turns out to be the weak part of our overall VQA approach.  By using the scene graph generation procedure proposed in \cite{yang2018scene}, we found that in the cases where the answer to an open question was contained in the scene graph, our method was able to achieve  $53.24\%$ accuracy on this subset of the data (which could be compared to the 50.22\% accuracy of TRRNet). We are currently working on improving the scene graph generation framework by integrating recent advancements in object detection such as \cite{tan2019efficientdet}  or in scene graph generation \cite{zellers2018neural,zhang2019graphical,koner2020relation}. We hope that these methods lead to more accurate scene graphs so that our method is able to retain close to human performance as presented in this paper.


\if false
In addition, we experimented with modification to MODELNAME. Initially, we conducted experiments where the agent's starting location on the scene graph was chosen uniformly at random. Then we ran multiple rollouts with different starting nodes and a joint prediction was formed via a majority vote. However, this procedure is problematic when the scene graph consists of multiple connected components. Therefore, we introduced an artificial hub node as starting location that allows the agent to transition to any other node in the graph. Initially we had concerns towards this approach because the node embedding formed by the GAT are expressive enough so that the agent can learn to make one transition to the answer node and remain there. This would undermine the philosophy of our multi-hop reasoning approach and limit the interpretability. However, we found that the agent still explores the scene graphs in this setting. In many cases, the agent even makes the initial transition to the correct answer node but keeps exploring the neighborhood in order to extract relevant information before returning to the answer node. An example for this observation is shown in Figure \ref{subfig:VQA_ex1} and \ref{subfig:VQA_ex2}.
\fi

\section{Conclusion}
\label{sec:conclusion}
We have proposed a novel method for visual question answering based on multi-hop sequential reasoning and deep reinforcement learning. Concretely, an agent is trained to extract conclusive reasoning paths from scene graphs. 
To analyze the reasoning abilities of our method in a controlled setting, we conducted a preliminary experimental study on manually curated scene graphs and concluded that our method reaches human performance. In future works, we plan to  incorporate state-of-the-art scene graph generation into our method to cover the complete VQA pipeline.


\newpage
\bibliography{bibliography}
\bibliographystyle{icml2020}

\end{document}